\documentclass[runningheads]{llncs}
\usepackage[mobile]{eccv}
\usepackage{eccvabbrv}

\usepackage[accsupp]{axessibility}
\usepackage{booktabs}
\usepackage{colortbl}
\usepackage{diagbox}
\usepackage{graphicx}
\usepackage{hhline}
\usepackage{makecell}
\usepackage{mathtools}
\usepackage{xspace}
\usepackage{pifont}
\newcommand{\cmark}{\ding{51}}%

\usepackage{hyperref}
\usepackage{orcidlink}
\usepackage{multirow}
\newcolumntype{P}[1]{>{\centering\arraybackslash}p{#1}}

\DeclarePairedDelimiter\floor{\lfloor}{\rfloor}
\DeclarePairedDelimiter\ceil{\lceil}{\rceil}
\definecolor{LightCyan}{rgb}{1., 0.83529412, 0.50196078}

\newcommand{\method}{N2F2\xspace}
\newcommand{\methodfull}{Nested Neural Feature Fields\xspace}

\makeatletter
\renewcommand{\paragraph}{%
  \@startsection{paragraph}{4}%
  {\z@}{0.5em}{-0.5em}%
  {\normalfont\normalsize\bfseries}%
}
\makeatother

\title{\method: Hierarchical Scene Understanding with Nested Neural Feature Fields}
\titlerunning{Nested Neural Feature Fields}
\author{Yash Bhalgat\orcidlink{0000-0001-7775-6250} 
\quad
Iro Laina 
\quad
Jo\~{a}o F.\ Henriques 
\\
Andrew Zisserman
\quad
Andrea Vedaldi
}
\authorrunning{Y.~Bhalgat \etal}
\institute{Visual Geometry Group, University of Oxford
\email{\{yashsb,iro,joao,az,vedaldi\}@robots.ox.ac.uk}}

\begin{document}
\maketitle
\begin{abstract}
Understanding complex scenes at multiple levels of abstraction remains a formidable challenge in computer vision.
To address this, we introduce Nested Neural Feature Fields (\method), a novel approach that employs hierarchical supervision to learn a \textit{single} feature field, wherein different dimensions within the same high-dimensional feature encode scene properties at varying granularities.
Our method allows for a flexible definition of hierarchies, tailored to either the physical dimensions or semantics or \textit{both}, thereby enabling a comprehensive and nuanced understanding of scenes.
We leverage a 2D class-agnostic segmentation model to provide semantically meaningful pixel groupings at arbitrary scales in the image space, and query the CLIP vision-encoder to obtain language-aligned embeddings for each of these segments.
Our proposed hierarchical supervision method then assigns different nested dimensions of the feature field to distill the CLIP embeddings using deferred volumetric rendering at varying physical scales, creating a coarse-to-fine representation.
Extensive experiments show that our approach outperforms the state-of-the-art feature field distillation methods on tasks such as open-vocabulary 3D segmentation
and localization,
demonstrating the effectiveness of the learned nested feature field.
\keywords{Hierarchical Scene Understanding \and Feature Field Distillation \and Open-Vocabulary 3D Segmentation}
\end{abstract}

\section{Introduction}%
\label{sec:intro}

\begin{figure}[t]
\centering
\includegraphics[width=\textwidth,trim={0.5cm 5.7cm 1.5cm 1cm},clip]{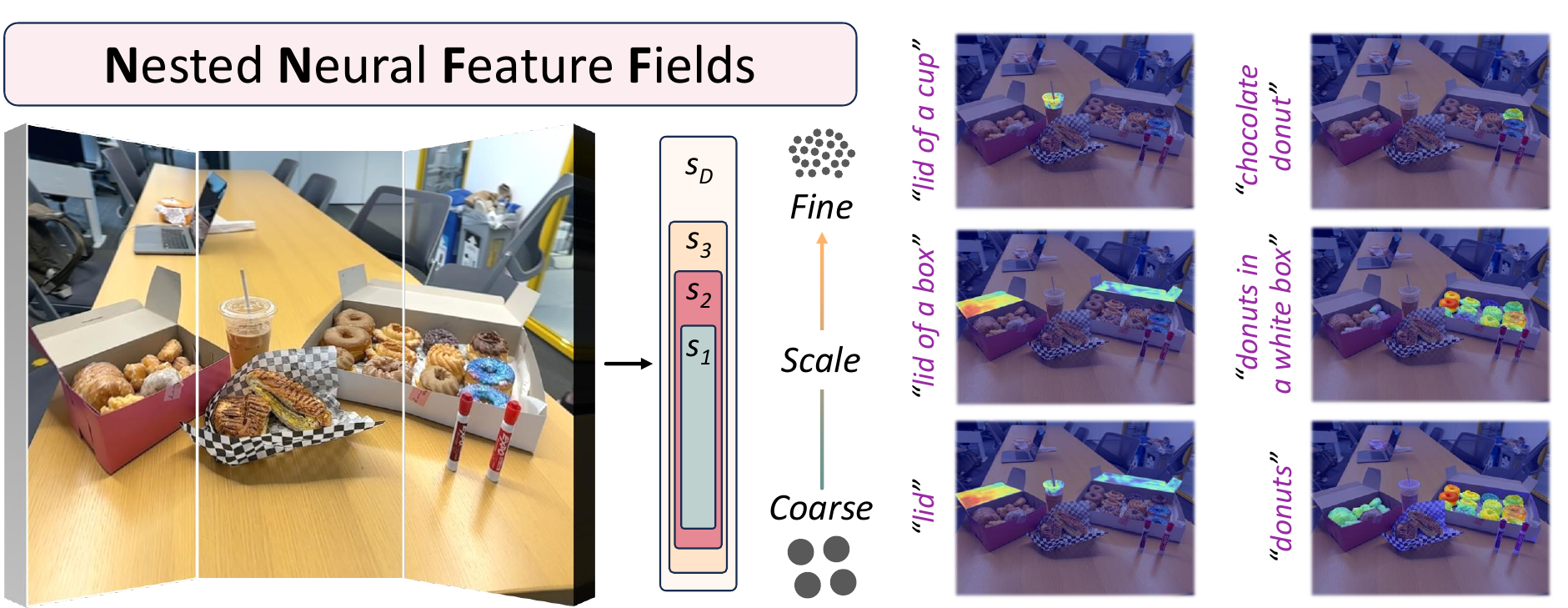}
\caption{\textbf{Nested Neural Feature Fields (N2F2)}. We present N2F2, wherein different dimensions of the same feature field encode scene properties at varying granularities.
The illustration captures the essence of hierarchical scene understanding, depicting how our model differentiates between \textit{coarse} and \textit{fine} scales to accurately interpret complex semantic queries, such as \textit{``donuts in a white box''} and \textit{``chocolate donut''}, showcasing the model's versatility in handling detailed object descriptions within 3D environments.}%
\label{fig:teaser}
\end{figure}

3D scene understanding is an important problem in computer vision which still presents several challenges.
One of them is that scene understanding is inherently hierarchical, as it requires reasoning about the scene at varying levels of geometric and semantic granularity.
Models must simultaneously understand the high-level structure and composition of the scene as well as fine-grained object details.
This is important for applications like robotics and augmented reality.

Recent progress in radiance fields has played a pivotal role in advancing 3D scene understanding.
Methods such as Neural Radiance Fields (NeRF)~\cite{mildenhall20nerf:} and 3D Gaussian Splatting~\cite{kerbl233d-gaussian} can extract the shape and appearance of 3D scenes without 3D supervision or specialized sensors, as they are inferred directly from several RGB images via differentiable rendering.
Furthermore, this optimization process can be extended to distill and fuse 2D information into the 3D representation well beyond RGB values.
Several authors have in fact proposed \emph{fusion approaches}, where 2D labels~\cite{zhi21in-place, bhalgat23contrastive, kim24garfield:, siddiqui23panoptic} or 2D features~\cite{tschernezki22neural,kobayashi22decomposing, kerr2023lerf} are extracted from multiple views of the scene and fused into a single 3D model.
The fused features not only augment the 3D reconstruction with a semantic interpretation, but can also remove noise from the 2D labels, improving their quality.

In this work, we focus on the problem of 3D distillation of vision-language representations, such as CLIP~\cite{radford21learning}, which, in turn, enables open-vocabulary 3D segmentation and localization of objects or scene elements based on natural language descriptions.
Existing methods like LERF~\cite{kerr2023lerf} have demonstrated the potential of embedding language features into NeRFs, allowing users to query 3D scenes with arbitrary text inputs, but face two key limitations.

First, while vision-language models exhibit remarkable few-shot transfer capabilities, their performance often degrades for more complex linguistic constructs like compound nouns (``paper napkin'') or partitive phrases (``bag of cookies''). This is because individual components may be interpreted separately (\eg, causing to detect ``paper'' and ``napkin'' as separate concepts instead of a ``paper napkin'').
This stems from the inherent challenge of compositional generalization that plagues such models and causes them to behave like bags-of-words~\cite{yuksekgonul2022and, thrush2022winoground, ma2023crepe, lin2023visualgptscore}.
This behavior makes it difficult to use vision-language models to capture compositions of objects and their attributes or relations and is often attributed to the text encoder bottleneck~\cite{kamath2023text} and the contrastive formulation that is used to train such models.
In practice, this means that querying CLIP with the aforementioned prompts would falsely localize both paper \emph{and} napkins, bags \emph{and} cookies.
Existing 3D feature distillation methods directly inherit these shortcomings, failing to accurately localize or segment objects described by compound expressions.

A second limitation of methods like LERF is their inefficiency during inference.
To produce relevance maps for a given text query, these approaches densely evaluate the feature field at multiple spatial scales.
This imposes a noticeable computational burden that grows linearly with the number of scales processed.

This work aims to address both limitations.
Our key insight is to impose a \emph{hierarchical structure} on the 3D feature field during training.
Specifically, we propose \textbf{\methodfull (\method)}, wherein different subsets of dimensions within a single high-dimensional feature field are tasked with encoding scene properties at varying granularities.
This design choice enables \method to simultaneously capture multi-scale scene representations in a coherent and parameter-efficient manner.
To address the compositionality challenge, we further contribute a novel composite embedding approach that, during inference, combines the features across all hierarchy levels in a weighted manner, effectively aggregating relevance scores across multiple scales given a text query and eliminating the need for explicit scale selection.

Through extensive experimentation on challenging datasets, we demonstrate that \method significantly outperforms prior work on open-vocabulary 3D segmentation and localization, including those involving complex compound queries.
Moreover, our composite embedding strategy yields considerable speedups during inference, making N2F2 1.7$\times$ faster than the current leading approach, LangSplat~\cite{qin2023langsplat}, with better accuracy and a significant increase in granularity.

In summary, our core contributions are:
(1) A hierarchical supervision framework for distilling multi-scale semantic representations into a unified 3D feature field;
(2) a composite embedding strategy that enables efficient open-vocabulary querying without explicit scale selection;
(3) state-of-the-art performance on challenging open-vocabulary 3D segmentation and localization benchmarks, with particular gains on complex compound queries.

\section{Related Work}

\paragraph{Radiance Fields.}

Radiance fields have emerged as a powerful representation for capturing and rendering complex 3D scenes from a set of multi-view posed images.
Neural Radiance Fields (NeRFs)~\cite{mildenhall20nerf:} pioneered this approach by representing a scene as a continuous 5D function, mapping spatial coordinates and viewing directions to colors and densities.
These representations can be optimized by leveraging differentiable volume rendering.
This seminal work has inspired a plethora of variants and extensions, such as deformable or dynamic NeRFs~\cite{pumarola20d-nerf:}, NeRFs for unconstrained images~\cite{martin-brualla21nerf}, and variants for improved efficiency~\cite{chen22tensorf:,chan22efficient,sun22direct}.
Recent work on 3D Gaussian Splatting (3DGS)~\cite{kerbl233d-gaussian} offers an alternative by representing scenes as a mixture of 3D Gaussians, which has  shown impressive quality and speed in 3D reconstruction.
The real-time rendering of 3DGS has enabled various downstream applications in dynamic scene reconstruction~\cite{wu234d-gaussian}, editing~\cite{fang23gaussianeditor:,chen23gaussianeditor:} as well as generation~\cite{tang23dreamgaussian:}.

\paragraph{3D Scene Segmentation.}

Motivated by the success of radiance fields, several studies have successfully extended such representations to 3D segmentation models.
Examples include semantic segmentation~\cite{zhi21in-place,vora21nesf:}, panoptic segmentation~\cite{bhalgat23contrastive,siddiqui23panoptic,fu22panoptic, liu2023instance} and part segmentation~\cite{zarzar2022segnerf}.
A common characteristic of these works is that, given multi-view images of a scene and corresponding 2D labels, these labels are fused into the 3D space as part of the radiance field optimization process.
While NeRF has gained significant attention, it is worth noting that the concept of integrating multiple viewpoints and semantic information for 3D reconstruction predates NeRF and has been explored in various methods~\cite{hermans2014dense,mccormac2017semanticfusion, sunderhauf2017meaningful, ma17multi-view, mascaro21diffuser:,vineet2015incremental}.

More recently, several authors have turned to the Segment Anything model (SAM)~\cite{kirillov2023segment} to obtain class-agnostic 3D segmentations~\cite{ying2024omniseg3d, kim24garfield:, cen2024segment, cen2023segment,hu2024semantic}.
SAM offers a promptable approach to segmentation with multiple levels of granularity, which has led most of these studies to adopt an interactive procedure, based on user-provided object seeds.
A notable exception is GARField~\cite{kim24garfield:}, which operates independently of user input.
Instead, it leverages masks produced by SAM, associating each mask with its corresponding 3D scale, and optimizes a scale-conditioned affinity field.
GARField can either be interactively queried at specific scales or produce a hierarchy of groupings automatically.
However, it does not offer an automatic scale selection mechanism.

\paragraph{3D Feature Distillation.}

Another related line of work focuses on lifting 2D image features to the 3D space~\cite{tschernezki22neural, kobayashi22decomposing, zhou2023feature, goel2023interactive}.
Similarly to the segmentation case, distilling features from a 2D teacher model, such as DINO~\cite{caron21emerging}, amounts to optimizing a 3D feature field (alongside the radiance field) to reconstruct the teacher features.
In such cases, 3D segmentation involves computing the similarity of the features embedded in the 3D feature field to some query.

Closest to our work are methods that construct 3D language fields from image-text features~\cite{kobayashi22decomposing, kerr2023lerf, zuo2024fmgs, qin2023langsplat, liu2023weakly, chen2024panoptic,engelmann2023opennerf}, which then enables querying the 3D representation with open-vocabulary text descriptions, thus achieving text-guided segmentation.
In particular, LERF~\cite{kerr2023lerf} is a scaled-conditioned approach that distills multi-scale CLIP~\cite{radford21learning} features into a NeRF model.
However, its segmentation quality is limited since it does not use mask supervision and relies mainly on the CLIP encoder which produces global features that lack precise localization.
LangSplat~\cite{qin2023langsplat} addresses this issue by combining CLIP features and SAM masks in a 3DGS representation.
Unlike LERF or GARfield, it only comprises three distinct scales (the same as SAM), which improves the efficiency of the method but reduces the granularity of the semantic hierarchy.

During inference, both LERF and LangSplat need to densely evaluate the feature field at multiple spatial scales to select the best scale for a given query.
This results in a noticeable overhead, especially as the number of scales increases (\eg, as in LERF\@).
Our \method approach makes use of the same underlying 2D models but is simultaneously \textit{more granular} than LangSplat and \textit{more efficient} than both methods while obviating the need for explicit scale selection.

Beyond radiance fields, several works~\cite{peng2023openscene,ding2023pla,ha2022semantic,zhang2023clip,jatavallabhula2023conceptfusion} operate on point-cloud data and align dense point-features with text and/or image pixels using pre-trained 2D vision-language models.

\paragraph{Hierarchical Representation Learning}

focuses on developing multi-layered data abstractions to enhance model adaptability and interpretability across tasks. 
Matryoshka representation learning (MRL)~\cite{kusupati2022matryoshka} aims to embed multiple levels of granularity into a single representation allowing the learned embeddings to adapt to the varying computational constraints of the task at hand.
MERU~\cite{desai2023hyperbolic} presents a contrastive approach that yields hyperbolic representations of images and text, resulting in interpretable and structured representations while being competitive with CLIP embeddings on various downstream tasks.
In the domain of NLP, TreeFormers~\cite{patel2022forming} introduce a general-purpose text encoder that learns a composition
operator and pooling function to construct hierarchical encodings for natural language expressions leading to improvements in compositional generalization.
Our approach, inspired by MRL, aims to learn a single feature field wherein different subsets of the feature field encode varying levels of scene properties based on both physical scale and the underlying semantics.

\section{Method}%
\label{sec:method}

\begin{figure}[t]
\centering
\includegraphics[width=\textwidth]{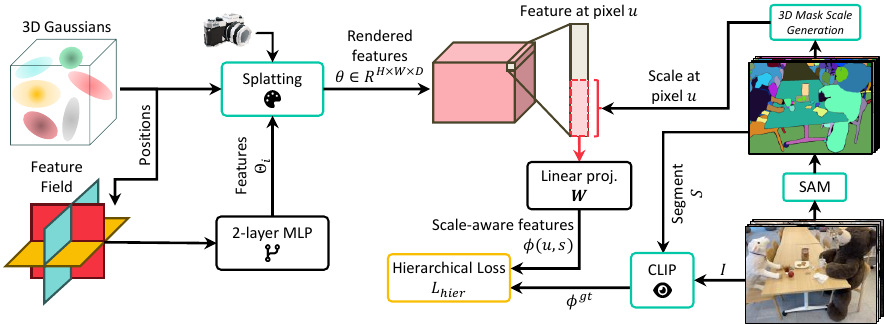}
\caption{\textbf{\method Overview.}
\textit{Left}:
\method employs 3D Gaussian Splatting (3DGS)  to represent the scene, augmented with a feature field that captures scene properties across different scales and semantic granularities.
\textit{Middle}:
Different subsets of the same feature vectors encode scene properties at varying scales.
This unified feature field is optimized using a hierarchical supervision loss applied to the scale-aware features.
\textit{Right}:
We extract a pool of segments using SAM and pre-compute a CLIP embedding for each.
Each segment is assigned a physical scale computed using the 3DGS model, which is then used to compute the scale-aware feature.
}%
\label{fig:arch}
\end{figure}

In this section, we describe \textbf{\methodfull (\method)}, an approach to learning multi-scale semantic representations for 3D scenes that uses a form of hierarchical supervision.
We first describe the underlying feature field architecture (\cref{sec:arch}).
Then, we introduce the proposed hierarchical supervision method (\cref{sec:hierarch}) which encodes varying levels of scene granularity into different subsets of the same feature space, leading to the idea of nested feature fields.
Finally, in \cref{sec:comp-emb}, we formulate a novel composite embedding approach that enables \method to handle compound open-vocabulary queries during inference.
\Cref{fig:arch} provides a high-level overview of \method.

\subsection{Feature Field Architecture}%
\label{sec:arch}

\subsubsection{3D Scene Representation.}

Given a set of images $\mathcal{I}$ of a scene with a corresponding camera pose $\pi \in SE(3)$ for every image $I \in \mathcal{I}$, we aim to construct a language-aware 3D model of the scene.
We use 3D Gaussian Splatting (3DGS)~\cite{kerbl233d-gaussian} as the scene representation, where the scene is modeled by a mixture of 3D Gaussians.
For any point $x\in\mathbb{R}^3$ in the scene, the influence function of the $i^{th}$ Gaussian is parametrized as:
\[
g_i(x) = \exp \left( -\frac{1}{2} (x - \mu_i)^\top \Sigma_i^{-1} (x - \mu_i) \right),
\]
where $\mu_i \in \mathbb{R}^3$ is the Gaussian mean or center and $\Sigma_i \in \mathbb{R}^{3\times3}$ is its covariance, expressed by a scaling matrix $S_i$ and rotation matrix $R_i$ as $\Sigma_i=R_i S_i S_i^\top R_i^\top$.
To model the 3D radiance field, each Gaussian also has an opacity $\sigma_i \in \mathbb{R}_+$ and a view-dependent colour $c_i(\nu)$ given by spherical harmonic coefficients $\mathcal{C}_i\in\mathbb{R}^k$ (up to the $3^{rd}$ order).
Overall, the 3DGS model is given by
$
G=\{(\mu_i,\Sigma_i,\sigma_i,\mathcal{C}_i)\}_i
$.
Given the model $G$, the 3D radiance field is defined as:
\begin{equation}
\sigma(x)
=
\sum_{i}
\sigma_i g_i(x),
~~~
c(x,\nu)
=
\frac{ \sum_{i} c_i(\nu) \sigma_i g_i(x)}{ \sum_{j} \sigma_j g_j(x)}.
\end{equation}
Given $G$ and a camera viewpoint $\pi$, the differentiable Gaussian Splatting renderer~\cite{kerbl233d-gaussian} produces an image $\hat{I}=\mathcal{R}(G,\pi)\in\mathbb{R}^{H \times W \times 3}$.
The key reason for choosing 3DGS as the scene representation rather than NeRF~\cite{mildenhall20nerf:} or its faster variants~\cite{alex-yu-and-sara-fridovich-keil21plenoxels:,muller22instant,chen22tensorf:,sun22direct} is that 3DGS provides a very efficient renderer (both in speed and memory) resulting in real-time rendering of full-sized images as well as feature maps, which allows us to perform rapid querying at inference time.

\subsubsection{Gaussian Feature Field.}

In this work, we propose to augment the 3DGS model with a feature field $\Theta: \mathbb{R}^3 \rightarrow \mathbb{R}^D$.
A natural choice for doing this would be to associate each Gaussian with a learnable feature vector and optimize these features in the same manner as the other parameters (\ie, $\mu_i,\Sigma_i,\sigma_i,\mathcal{C}_i$).
However, in practice, the size $D$ tends to very high (\eg, $D=512$ for CLIP embeddings) which drastically increases the number of parameters in the augmented 3DGS model.
To address this issue, we model the feature field $\Theta$ with a memory-efficient TriPlane representation~\cite{fridovich-keil23k-planes:} followed by a 3-layer MLP\@.
During rendering, we query this hybrid representation at the Gaussian centers to get the associated feature for each Gaussian, \ie, $\Theta_i=\Theta(\mu_i)$.%
\footnote{We use the same variable name $\Theta_i$ for simplicity.}
Thus, given a camera viewpoint $\pi$, this results in a rendered pixel-level feature map:
\begin{equation} \label{eq:feat-map}
    \theta=\mathcal{R}(\Theta,G,\pi)\in\mathbb{R}^{H\times W\times D}
\end{equation}
In addition, to save memory during training, we first render the TriPlane features, then use them as input to the 3-layer MLP to obtain the pixel/ray feature.
This deferred rendering is performed only during training and leads to a significantly faster optimization process without any loss in performance. Please refer to implementation details in the appendix for differences to test time.

\subsection{Scale-aware Hierarchical Supervision}%
\label{sec:hierarch}

Our main goal is to learn a unified representation that captures the meaning of the scene across different scales and semantic granularities.
We do so by modeling \method with a \emph{single} Gaussian feature field, where different feature dimensions represent varying levels of detail, from the overall scene structure to fine-grained object particulars.
To achieve this, we train our feature field model using a scale-aware hierarchical supervision method described below.

\subsubsection{Extracting Training Data.}

To get an accurate geometry model
for each scene, we first optimize the radiance field related parameters of the 3DGS model. Then, we use SAM~\cite{kirillov2023segment} to extract class-agnostic but semantically meaningful segments from every image $I \in \mathcal{I}$, resulting in a pool of segments $\mathcal{S}$ spanning all the images.
Following GARField~\cite{kim24garfield:}, for each segment $S\in\mathcal{S}$, we use the expected depth values from the 3DGS model to lift the pixels in the segment to a corresponding 3D point cloud.
We obtain the scale of this point cloud as the largest eigenvalue of the covariance matrix of the 3D point positions.
We then quantize the extracted scales into $D$ bins using a quantile transformation, where $D$ is the dimension of the feature vectors in our field.\footnote{This transforms the scales to follow a uniform distribution and also reduces the impact of marginal outliers.}
We denote the quantized scale corresponding to segment $S$ as $s \in [0,1)$.
Next, for every segment $S\in\mathcal{S}$ of an image $I$, we use the CLIP~\cite{radford2021learning} image encoder to obtain a language embedding $\phi^{\operatorname{gt}} \in \mathbb{R}^{D}$:
\begin{equation}\label{eq:lang-gt}
    \phi^{\operatorname{gt}} = E(I \odot S),
\end{equation}
where $E$ is the CLIP vision encoder and $\odot$ is the Hadamard product.

We use this vector as ``ground truth'' to optimize our \method model with hierarchical supervision. 
To do this, we associate each dimension of the feature field $\Theta \in \mathbb{R}^{D}$ to a quantized scale value (note that the dimension $D$ is also the number of quantized values), such that the lower dimensions are mapped to larger (coarser) scales and higher dimensions to finer scales.
This is predicated on the observation that objects or scene components at larger scales can be differentiated with fewer dimensions, using less specific features.
Conversely, as one zooms into finer scales, more dimensions become necessary to match the increased specificity and diversity of the features.
Mathematically, for a quantized scale $s \in [0,1)$, we associate the dimension $M(s)$ defined by the mapping $M(s)=\ceil{D\cdot(1-s)}$, where $\ceil{\cdot}$ is the ceiling function.

\subsubsection{Scale-aware Feature.}

Consider a pixel $u$ sampled from a segment $S$ with scale $s$.
The rendered feature $\theta(u)$ can be obtained from our feature field using \cref{eq:feat-map}.
Now, given the mapping $M(\cdot)$, we obtain the \textit{scale-aware} feature as:
\begin{equation}\label{eq:proj}
    \phi(u,s) = \mathbf{W}_{1:M(s)}\theta(u)_{1:M(s)}
\end{equation}
Here, $\mathbf{W}\in\mathbb{R}^{D \times  D}$ is a learnable projection matrix.
$\mathbf{W}_{1:M(s)}\in \mathbb{R}^{D\times M(s)}$ is formed by the first $M(s)$ columns of $\mathbf{W}$ and $\theta(u)_{1:M(s)}$ are the first $M(s)$ elements of the feature $\theta(u)$.
Hence, the resulting scale-aware feature $\phi(u,s)\in\mathbb{R}^D$.
We do not explicitly normalize $\phi$ or $\theta$ so that the former remains a linear function of the latter.

Intuitively, the projection matrix $\mathbf{W}_{1:M(s)}$ dynamically adjusts to the variable dimensionality of the feature vector $\theta(u)_{1:M(s)}$ corresponding to pixel $u$ at scale $s$. It ensures that this vector is mapped to a fixed dimensionality $D$ that matches that of the features $\phi^{\operatorname{gt}}$.
This enables the use of \textit{single-scale} features to supervise the training of a nested \textit{multi-scale} feature field.

\subsubsection{Hierarchical Loss Supervision.}

During training, we sample pixels $u$ uniformly across the image set $\mathcal{I}$.
Since there can be multiple segments associated with the same pixel, we sample a segment $S$ with a probability inversely proportional to the $\log$ of the area of $S$.
With the rendered features $\theta(u)$ and the language-aligned teacher embeddings $\phi^{\operatorname{gt}}$ from \cref{eq:lang-gt}, we minimize the loss:
\begin{equation}\label{eq:loss-hier}
    \mathcal{L}_{\text{hier}} = \sum_{u,s} \mathcal{L}(\phi(u,s),\phi^{\operatorname{gt}})
\end{equation}
where $\mathcal{L}=\mathcal{L}_{\text{2}}+\lambda\mathcal{L}_{\text{cos}}$ is a weighted combination of the L2 and cosine distances.

\subsection{Composite Embedding for Open-Vocabulary Querying}%
\label{sec:comp-emb}

Once the model has been optimized with the hierarchical loss function, it can be queried with language prompts encoded using the CLIP text encoder.
The querying can be done either in 3D with the point features at the Gaussian centers or in 2D with the rendered 2D features.
We discuss the latter in this section, but our method is compatible with both scenarios.

Previous methods~\cite{kerr2023lerf,qin2023langsplat}, compute relevancy maps at multiple scales (\eg, LERF uses 30 scales) for each text query and then choose the scale with the highest relevancy score.
We notice that this leads to a noticeable overhead for every text input, especially for a large number of scales.
Hence, we propose a \textit{composite embedding} method that allows us to compute only \textit{one} relevancy map per text query and use that as the output.

We define the \textit{composite embedding} at a pixel $u$ to be a weighted sum of all possible scale-aware features at that pixel.
The idea of using a linear combination of features across scales arises from the necessity to handle compound queries effectively.
Consider the query ``lid of a cup''.
The scale-aware features ideally trigger twice for pixels that are both on a ``lid'' and on a ``cup'', akin to an intersection of sets.
This is because the composite embedding aggregates features from the scales corresponding to ``a cup'' and ``a lid'', effectively highlighting the summed response of these features.
Conversely, querying ``a cup'' would trigger once for all pixels on a cup, reflecting the broader, single-scale interpretation.

The weights in the linear combination are \textit{query-agnostic} and account for the relevance of each scale
for \textit{any} given query.
Their goal is to select scales that are more strongly related to specific types of queries, and are obtained by comparing each scale-specific feature vector to prototypical phrases such as ``\textit{object}'', ``\textit{stuff}'', ``\textit{thing}'', ``\textit{part}'' and ``\textit{texture}''.%
\footnote{
LERF~\cite{kerr2023lerf} uses similar phrases to compute relevancy scores \textit{per query}.}
Mathematically, given a point feature for a Gaussian, $\Theta_i$, the weight $\gamma^\text{3D}_i$ is:
\begin{equation} \label{eq:gamma-3D}
    \gamma^\text{3D}_i = \underset{d}{\text{Softmax}}
    \left(
        \max_k \,(\mathbf{W}_{1:d} \Theta_{i,1:d})^\top \phi^{\text{canon}}_k
    \right),
\end{equation}
where $\{\phi^{\text{canon}}_k\}$ is the set of CLIP embeddings of the predefined canonical phrases listed above.
The $\gamma^\text{3D}$ tensor only needs to be computed once post-training. For a new viewpoint, $\gamma^\text{3D}$ is rendered to obtain the pixel-level tensor $\gamma\in\mathbb{R}^{H\times W\times D}$.

Since there are exactly $D$ quantized scales, each mapped to a dimension of the feature field, the composite embedding is $\phi_{\text{comp}}(u) = \sum_{d=1}^{D} \gamma_d(u) \phi(u,s_d)$.
Expanding this using \cref{eq:proj}, we get:
\begin{align}
\phi_{\text{comp}}(u) &= \sum_{d=1}^{D} \gamma_d(u) \mathbf{W}_{1:d} \theta(u)_{1:d}
                   = \sum_{d=1}^{D} \gamma_d(u) \left(\sum_{j=1}^{d} \theta(u)_j W_j\right) \nonumber \\
                   &= \sum_{j=1}^{D} \left(\sum_{d=j}^{D} \gamma_d(u)\right) \theta(u)_j W_j \nonumber
\end{align}
We can rewrite this as $\phi_{\text{comp}}(u) = \mathbf{W} \tilde{\theta}(u)$, where
\begin{equation} \label{eq:gamma}
\tilde{\theta}(u) = \begin{bmatrix}
\sum_{d=1}^{D} \gamma_d(u) \\
\sum_{d=2}^{D} \gamma_d(u) \\
\vdots \\
\gamma_D(u)
\end{bmatrix} \odot \theta(u) , \quad \text{or equivalently,} \quad
\tilde{\Theta}_i = \begin{bmatrix}
\sum_{d=1}^{D} \gamma_{i,d}^\text{3D} \\
\sum_{d=2}^{D} \gamma_{i,d}^\text{3D} \\
\vdots \\
\gamma_{i,D}^\text{3D}
\end{bmatrix} \odot \Theta_i
\end{equation}
Note, $\gamma_{i,d}^\text{3D}$ denotes the $d^{th}$ element of the vector $\gamma_{i}^\text{3D}$ corresponding to the $i^{th}$ Gaussian.
Given the above formulation, instead of computing the point-features $\Theta_i$, we pre-compute the point-features $\tilde{\Theta}_i$ prior to the text-querying process.
\section{Experiments}%
\label{s:experiments}

\subsubsection{Tasks and Datasets.}

We assess the ability of \method to perform open-vocabulary \textit{segmentation} and \textit{localization} on challenging natural language queries.
We train and evaluate our method on the expanded LERF dataset made available by Qin \etal~\cite{qin2023langsplat}.
This expanded version contains ground-truth segmentation masks for textual queries required to evaluate the segmentation performance.
The expanded version also contains additional and more challenging localization samples to evaluate the localization accuracy.
We also use the 3D-OVS dataset~\cite{liu2023weakly} to evaluate and compare our open-vocabulary 3D segmentation capabilities.

\subsubsection{Metrics.}

For the 3D localization task, we consider a text query to be successfully localized if the identified highest-relevancy pixel in a view lies inside the ground-truth bounding box, reporting the localization accuracy.
We report the mIoU scores for the 3D segmentation task.

\subsubsection{Baselines.}

We compare our method with LERF~\cite{kerr2023lerf} and LangSplat~\cite{qin2023langsplat}, which are the current state-of-the-art feature distillation methods for open-vocabulary 3D segmentation and localization.
We also compare with a baseline distilling the pixel-aligned feature from LSeg~\cite{li2022languagedriven} into 3D. Note that, both LangSplat and our method utilize SAM~\cite{kirillov2023segment} during optimization.

\subsection{Results}

\begin{table}[t]
  \renewcommand\tabcolsep{3pt}
  \centering
  \caption{Localization accuracy (\%) comparisons on LERF scenes, averaged across text queries.
  More query-specific results are provided in the appendix. The second column indicates if a method utilizes segmentation from SAM \cite{kirillov2023segment} during training.}
  \begin{tabular}{lc @{\hspace{14pt}} cccccc}
    \toprule
    Method                            & SAM & \textit{bouquet} & \textit{ramen} & \textit{figurines} & \textit{teatime} & \textit{waldo\_kitchen} & Overall       \\
    \midrule
    LSeg~\cite{li2022languagedriven} & & 50.0             & 14.1           & 8.9                & 33.9             & 27.3                    & 21.1          \\
    LERF~\cite{kerr2023lerf}         & & \textbf{91.7}    & 62.0           & 75.0               & 84.8             & 72.7                    & 73.6          \\
    LangSplat~\cite{qin2023langsplat} & \cmark & ---              & 73.2           & 80.4               & 88.1             & \textbf{95.5}           & 84.3          \\
    \midrule
    N2F2 (\textbf{Ours})              & \cmark & \textbf{91.7}    & \textbf{78.8}  & \textbf{85.7}      & \textbf{91.5}    & \textbf{95.5}           & \textbf{88.6} \\
    \bottomrule
  \end{tabular}%
  \label{table:lerf_loca}
\end{table}

\begin{table}[t]
  \renewcommand\tabcolsep{4pt}
  \centering
  \caption{Open-vocabulary 3D semantic segmentation performance (mIoU) on the expanded LERF dataset from Qin \etal~\cite{qin2023langsplat}, averaged across queries. The second column indicates if a method utilizes segmentation from SAM \cite{kirillov2023segment} during training.}%
  \label{table:lerf_seg}
  \begin{tabular}{lc @{\hspace{20pt}} ccccc}
    \toprule
    Method & SAM & \textit{ramen} & \textit{figurines} & \textit{teatime} & \textit{waldo\_kitchen} & Overall \\
    \midrule
    LSeg~\cite{li2022languagedriven}      & & 7.0   & 7.6       & 21.7    & 29.9          & 16.6    \\
    LERF~\cite{kerr2023lerf}             &  & 28.2  & 38.6      & 45.0    & 37.9          & 37.4    \\
    LangSplat~\cite{qin2023langsplat}    & \cmark  & 51.2  & 44.7      & 65.1    & 44.5          & 51.4    \\
    \midrule
    N2F2 (\textbf{Ours})                 & \cmark  & \textbf{56.6} & \textbf{47.0} & \textbf{69.2} & \textbf{47.9} & \textbf{54.4} \\
    \bottomrule
  \end{tabular}
\end{table}

\begin{figure}[t]
    \centering
    \includegraphics[width=\textwidth]{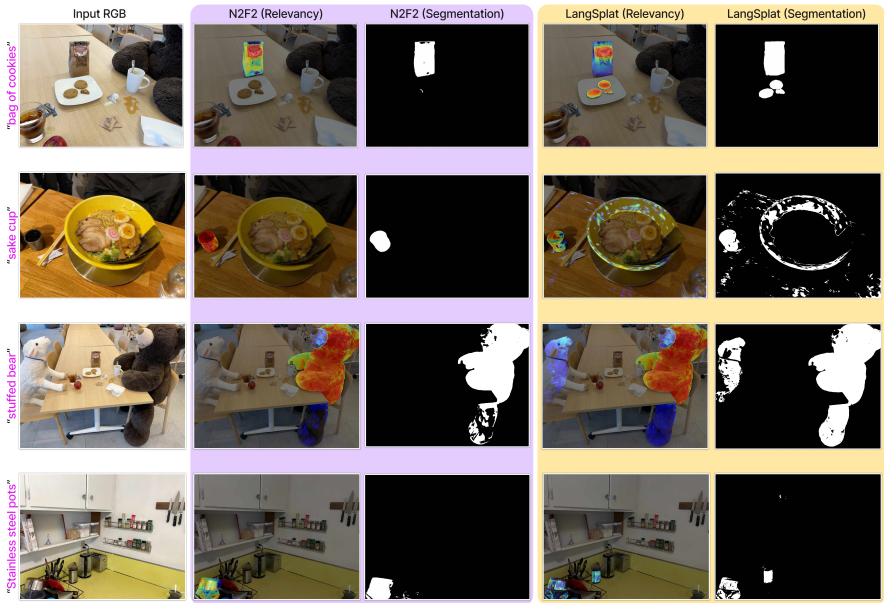}
    \caption{Qualitative comparisons with LangSplat~\cite{qin2023langsplat} on challenging \textit{compound} queries.}%
    \label{fig:enter-label1}
\end{figure}

\begin{table}[t]
  \renewcommand\tabcolsep{4pt}
  \centering
  \caption{Open-vocabulary 3D semantic segmentation performance (mIoU) on the 3D-OVS dataset~\cite{liu2023weakly}, averaged across queries. The second column indicates if a method utilizes segmentation from SAM \cite{kirillov2023segment} during training.}%
  \label{table:3dovs}
  \begin{tabular}{lc@{\hspace{20pt}}cccccc}
    \toprule
    Method           & SAM                   & \textit{bed}  & \textit{bench} & \textit{room} & \textit{sofa} & \textit{lawn} & Overall       \\
    \midrule
    LSeg~\cite{li2022languagedriven}    & & 56.0          & 6.0            & 19.2          & 4.5           & 17.5          & 20.6          \\
    ODISE~\cite{xu2023open}           &  & 52.6          & 24.1           & 52.5          & 48.3          & 39.8          & 43.5          \\
    OV-Seg~\cite{liang2023open}       &  & 79.8          & 88.9           & 71.4          & 66.1          & 81.2          & 77.5          \\
    \midrule
    FFD~\cite{kobayashi2022decomposing} & & 56.6          & 6.1            & 25.1          & 3.7           & 42.9          & 26.9          \\
    LERF~\cite{kerr2023lerf}          &  & 73.5          & 53.2           & 46.6          & 27            & 73.7          & 54.8          \\
    3D-OVS~\cite{liu2023weakly}       &  & 89.5          & 89.3           & 92.8          & 74            & 88.2          & 86.8          \\
    LangSplat~\cite{qin2023langsplat}  & \cmark & 92.5          & \textbf{94.2}  & \textbf{94.1} & 90.0          & 96.1          & 93.4          \\
    \midrule
    N2F2 (\textbf{Ours})             & \cmark   & \textbf{93.8} & 92.6           & 93.5          & \textbf{92.1} & \textbf{96.3} & \textbf{93.9} \\
    \bottomrule
  \end{tabular}
\end{table}

\subsubsection{3D Localization.}

\Cref{table:lerf_loca} shows the 3D localization performance on the expanded LERF dataset~\cite{qin2023langsplat}, on a diverse set of scenes, namely \textit{bouquet}, \textit{ramen}, \textit{figurines}, \textit{teatime}, and \textit{waldo\_kitchen}.
Notably, the \textit{bouquet} scene is not a part of the extended LERF dataset, and LangSplat~\cite{qin2023langsplat} does not provide results for this scene.
Thus, for this scene, we evaluate and compare on the text queries from the original LERF dataset.
Our method \method significantly outperforms the existing state-of-the-art methods, LERF~\cite{kerr2023lerf} and LangSplat~\cite{qin2023langsplat}, across most of the evaluated scenes, achieving an overall accuracy of $88.6\%$.
Our \method approach particularly outshines prior work in samples with complex \textit{compound queries}, such as ``sake cup'', ``bag of cookies'', etc. 
For a detailed breakdown of the performance across such queries, please refer to \Cref{sec:perf-comp}.

\subsubsection{3D Segmentation.}

\Cref{table:lerf_seg,table:3dovs} demonstrate the performance on the open-vocabulary 3D segmentation task on the expanded LERF~\cite{qin2023langsplat} and 3D-OVS~\cite{liu2023weakly} datasets respectively. The expanded LERF dataset contains various \textit{compound queries}, such as ``bag of cookies'', ``coffee mug'', and ``paper napkin'', in contrast to simple queries like ``cookies'', ``mug'', ``apple'', which makes it challenging to segment these referred objects.
LERF and LangSplat, which directly distill CLIP embeddings, struggle with such queries, producing a high relevancy score for all objects/attributes in the compound query.
An example shown in \cref{fig:enter-label1} shows that for ``bag of cookies'', LERF highlights both the ``bag'' \emph{and} the ``cookies'', although our method correctly highlights the ``bag'' only. 
More such examples are included in the appendix. Overall, our method achieves an mIoU of $54.5$ and outperforms both LERF and LangSplat on all scenes by a significant margin.

The 3D-OVS dataset contains relatively simple scenes where all included objects are of similar scales.
Thus, the performance of previous state-of-the-art methods is highly saturated with LangSplat achieving almost perfect segmentations for many text queries.
Nevertheless, our method outperforms LangSplat on $4$ out of $6$ scenes and improves the overall mIoU by $0.5\%$. 
More qualitative results are included in the appendix.

\subsection{Ablation Studies}

\subsubsection{Composite Embedding vs Explicit Scale Selection.}

Unlike LERF or LangSplat, our method does not explicitly perform scale selection for every query.
Instead, we design a composite embedding (and the $\gamma$ tensor, \cf \cref{eq:gamma}) to produce high relevancy at the appropriate scales and generate an aggregated relevancy map \textit{implicitly}.
But we could, in principle, also explicitly search the scale with the highest relevancy score in the same manner as LERF/LangSplat.
\Cref{tab:implicit-explicit} shows a comparison between the composite embedding and explicit scale selection 
for localization and segmentation on the expanded LERF dataset.
We observe that the composite embedding performs better than the explicit scale selection method while being approximately $5\times$ faster during querying.
Compared to \Cref{table:lerf_loca,table:lerf_seg}, we note that the \method explicit scale selection baseline \emph{also} outperforms LangSplat~\cite{qin2023langsplat}, which also uses explicit selection over 3 scales.
This further demonstrates the advantage of our hierarchical representation with increased granularity (\ie, $D$ scales).

Additionally, we ask --- \textit{what performance can \method achieve if the ideal scale was somehow provided?}
We call this the ``Oracle'' scale.
For a given text query, we define the oracle scale to be the scale that gives the highest performance for the task in question.
This gives an upper bound on the performance achievable by our method, which is shown in \cref{tab:implicit-explicit}.

\begin{table}[t] %
\centering
\setlength{\tabcolsep}{4pt}
\caption{Comparison of composite embedding based querying vs explicit scale-selection based querying on $4$ LERF scenes.
\textbf{Top}: 3D localization.
\textbf{Bottom}: 3D segmentation.}%
\label{tab:implicit-explicit}
\begin{tabular}{lccccc}
\toprule
Method                                         & \textit{ramen} & \textit{figurines} & \textit{teatime} & \textit{waldo\_kitchen} & Overall       \\
\midrule
\hspace{-2pt}\textit{\textbf{3D Localization}} &                &                    &                  &                         &               \\
Composite embedding                            & \textbf{78.8}  & \textbf{85.7}      & \textbf{91.5}    & \textbf{95.5}           & \textbf{87.9} \\
Explicit scale selection                       & \textbf{78.8}  & 83.9               & 89.8             & \textbf{95.5}           & 87.0          \\
\rowcolor{LightCyan}
\textit{Oracle} scale (Upper bound)            & 83.0           & 91.0               & 93.2             & 95.5                    & 90.6          \\
\hhline{======}
\noalign{\vspace{0.5ex}}
\hspace{-2pt}\textit{\textbf{3D Segmentation}} &                &                    &                  &                         &               \\
Composite embedding                            & \textbf{56.6}  & \textbf{47.0}      & \textbf{69.2}    & \textbf{47.9}           & \textbf{54.4} \\
Explicit scale selection                       & 55.7           & 45.9               & 66.8             & 46.3                    & 53.7          \\
\rowcolor{LightCyan}
\textit{Oracle} scale (Upper bound)            & 59.8           & 49.1               & 71.2             & 49.2                    & 57.3          \\
\bottomrule
\end{tabular}
\end{table}

\subsubsection{Efficiency.}

For each text query, LERF computes relevancy maps at $30$ scales (from $0$ to $2$) and LangSplat computes them at $3$ scales hardcoded to the SAM~\cite{kirillov2023segment} hierarchy.
For N2F2, with the composite embeddings, we can get the final output by computing only a single relevancy map per text query.
This leads to a significant speed up during querying making our method overall $1.7\times$ faster than LangSplat and $26\times$ faster than LERF, while using more scales.

\Cref{table:query-speed} shows a full breakdown of the querying time cost for each method.
With our method, we pre-compute $\gamma^{\operatorname{3D}}$ 
for every Gaussian in 3D.
Then, for each query, we render the $512$-dimensional composite embedding, in chunks.
Our method takes $0.1$s to produce a $730 \! \times \! 987 \! \times \! 512$ feature map from one viewpoint, while LangSplat takes $0.05$s to render and decode features to the same size.
Once rendered, LangSplat takes about $0.25$s/query, while our method takes about $0.16$s/query resulting in a $1.7$x speedup amortized over 10 queries per viewpoint.

\begin{table}[t]
  \renewcommand\tabcolsep{4pt}
  \centering
  \caption{Breakdown of inference-time querying speed for different methods.}
  \begin{tabular}{lccccc}
    \toprule
    Method           &  \makecell{Rendering\\time}  & \makecell{Time\\(per query)} & \makecell{Time\\(10 queries)} & \makecell{Total time\\(1 query)} & \makecell{Total time\\(10 queries)} \\
    \midrule
    LERF    & 18.4s &  2.5s &  20.0s  & 20.9s & 38.4s  \\
    LangSplat & 0.05s & 0.29s & 2.5s & 0.3s & 2.55s \\
    \method (\textbf{Ours})   &  0.1s & 0.16s &  1.4s & \textbf{0.24s} & \textbf{1.5s}  \\
    \bottomrule
  \end{tabular}%
  \label{table:query-speed}
\end{table}

\begin{table}[t]
  \renewcommand\tabcolsep{4pt}
  \centering
  \caption{Effect of step-size in the dimension-scale mapping. Evaluation on the LERF \textit{teatime} scene for the open-vocabulary 3D segmentation task (mIoU).}
  \begin{tabular}{lccccc}
    \toprule
    Scene           &  $k=1$ & $k=8$ & $k=32$ & $k=128$ & $k=170$ \\
    \midrule
    LERF (\textit{teatime}) & \textbf{69.2} & 68.7 & 68.1 & 67.2 & 66.4   \\
    \bottomrule
  \end{tabular}%
  \label{table:step-size}
\end{table}

\subsubsection{Effect of step-size in the dimension-scale mapping.}

In \cref{sec:hierarch}, we quantized the object scales to $D$ values, thereby associating each of the dimensions from $\{1,2,\dots,D\}$ with a quantized scale in the hierarchical loss optimization (\cf \cref{eq:loss-hier}).
We investigate what happens when we do not utilize all the dimensions for scale supervision.
That is, with a ``\textit{step-size}'' of $k$ in the dimension-scale mapping, we quantize the scales to $\floor{D/k}$ values and associate them to the dimensions $\{k,2k,\dots,k*\floor{D/k}\}$. For example, with $k=3$, only the dimensions $\{3,6,\dots,510\}$ are used.
We train models with $k=\{8,32,128,170\}$ using the hierarchical loss and compare them with the default model ($k=1$) on the LERF \textit{teatime} scene for open-vocabulary 3D segmentation.
\Cref{table:step-size} shows that performance gradually decreases as the granularity of the feature field is reduced, \ie, when we reduce the utilization of its dimensions.
We note that the case when $k=170$, \ie, when only $3$ different quantized scales or dimensions are used (namely $\{170,340,510\}$), is the closest to LangSplat which utilizes 3 scales.
In this setting, we achieve an mIoU of $66.4$ which is $1.3$ points higher than LangSplat's performance (\cf \cref{table:lerf_seg}).



\section{Limitations}

Despite the advancements \method brings to hierarchical scene understanding, it faces certain challenges. The model's ability to learn accurate feature fields heavily depends on the quality of scene reconstructions, necessitating a diverse array of images from varied viewpoints. While N2F2 accurately handles compound noun phrases (like ``coffee mug'') and partitive constructions (``bag of cookies'' or ``lid of the cup''), it struggles with global scene context queries, such as identifying a ``wooden desk in the corner of the room''. This limitation points to the challenge of integrating broad scene comprehension with specific object identification. 
Addressing these will be crucial for future developments in the field.

\section{Conclusion}

In this work, we introduced \methodfull (\method), a novel approach for hierarchical scene understanding.
N2F2 employs scale-aware hierarchical supervision to encode scene properties across multiple granularities within a unified feature field, significantly advancing open-vocabulary 3D segmentation and localization tasks.
Through extensive experiments, our approach demonstrated superior performance over state-of-the-art methods, such as LERF~\cite{kerr2023lerf} and LangSplat~\cite{qin2023langsplat}, highlighting the effectiveness of our hierarchical supervision methodology.
In particular, our method outshines on complex compounded and partitive constructions such as ``bag of cookies'', ``chair legs'', ``blueberry donuts'', etc.
Finally, we also propose a novel \textit{composite embedding} design that enables highly efficient querying as well as better performance than explicit scale-selection methods used in previous works.

\paragraph{Ethics.}

For further details on ethics, data protection, and copyright please see \url{https://www.robots.ox.ac.uk/~vedaldi/research/union/ethics.html}.

\section*{Acknowledgements} 
We are grateful for funding from EPSRC AIMS CDT EP/S024050/1 and AWS (Y.~Bhalgat), ERC-CoG UNION 101001212 (A.~Vedaldi and I.~Laina), EPSRC VisualAI EP/T028572/1 (I.~Laina, A.~Vedaldi and A.~Zisserman), and Royal Academy of Engineering RF\textbackslash 201819\textbackslash 18\textbackslash 163 (J.~Henriques).

\bibliographystyle{splncs04}
\bibliography{main,vedaldi_general,vedaldi_specific}


\appendix

\section{Performance on Compound Queries}%
\label{sec:perf-comp}

In Sec.\ 4.1 of the paper, we noted that, in comparison to previous methods (LERF~\cite{kerr2023lerf} and LangSplat~\cite{qin2023langsplat}), \method offers a strong advantage on queries which are \textit{compound} or \textit{partitive} in nature.
To understand this effect more clearly, we split the text queries from the expanded LERF dataset into two sets: (1) \textit{simple} single word queries such as ``cookies'', ``cup'', ``waldo'', ``spatula'', (2) \textit{compound or partitive} queries\footnote{For brevity, we refer to \textit{all} such queries as compound queries.} such as ``bag of cookies'', ``frog cup'', ``dark cup'', ``toy elephant'', ``toy cat statue''.
\Cref{tab:comp_labels} provides a full list of compound queries for each scene.

\Cref{tab:compound-results} shows that \method significantly outperforms LangSplat~\cite{qin2023langsplat}, the current state-of-the-art method, especially on compound queries where we see an improvement as big as \textbf{7.2} mIoU points on the \textit{ramen} scene.

To gain a clearer insight into which queries \method excels at, we rank the queries within each scene according to the performance disparity between \method and LangSplat.
We then highlight the top-3 and bottom-3 queries that exhibit the largest and smallest differences in performance, respectively.
As shown in \cref{tab:queries-sorted}, the largest performance improvements (\cf the `top-3 queries' column) are observed on compound queries and especially the ones which could apply to multiple objects, if they were not specific.
For example, the \textit{figurines} scene contains two similar looking rubber ducks, but only one of them wears a hat. Hence, the query ``rubber duck with hat'' must only segment the referred duck.
This is successfully achieved by \method but LangSplat segments parts of both the ducks resulting in a lower segmentation IoU.

\Cref{fig:teatime,fig:donuts,fig:figurines} demonstrate more qualitative comparisons between \method and LangSplat~\cite{qin2023langsplat} on various challenging \textit{compound} queries.

\begin{table}[t]
\renewcommand\tabcolsep{5pt}
\centering
\caption{Breakdown of performance on \textit{simple} and \textit{compound} queries for open-vocabulary 3D segmentation (mIoU). ``{comp.}'' is short for compound queries.}
\begin{tabular}{lcccccccc}
\toprule
\multirow{ 2}{*}{Method}                 & \multicolumn{2}{c}{\textit{ramen}} & \multicolumn{2}{c}{\textit{figurines}} & \multicolumn{2}{c}{\textit{teatime}} & \multicolumn{2}{c}{\textit{waldo\_kitchen}}\\
\cmidrule(l{4pt}r{4pt}){2-3} \cmidrule(l{4pt}r{4pt}){4-5} \cmidrule(l{4pt}r{4pt}){6-7} \cmidrule(l{4pt}r{4pt}){8-9}
                                         & simple                             & comp.                                  & simple                               & comp.                                      &  simple        & comp.         & simple        & comp.         \\
\midrule
LangSplat~\cite{qin2023langsplat}        & 65.4                               & 35.1                                   & 53.9                                 & 32.7                                       &  70.1          & 47.3          & 52.6          & 31.3          \\
N2F2 (\textbf{Ours})                     & \textbf{66.2}                      & \textbf{42.3}                          & \textbf{54.7}                        & \textbf{36.1}                              &  \textbf{72.2} & \textbf{50.9} & \textbf{53.2} & \textbf{35.5} \\
\bottomrule
\end{tabular}%
\label{tab:compound-results}
\end{table}

\begin{table}[t]
\renewcommand\tabcolsep{8.5pt}
\caption{Queries within each scene ranked according to performance disparity ($\Delta_{\text{perf}}$) between N2F2 and LangSplat~\cite{qin2023langsplat}.
Top-3 and bottom-3 ranked queries are shown.}%
\label{tab:queries-sorted}
\centering
\begin{tabular}{lcccc}
\toprule
Scene                                    & Top-3 queries                      & $\Delta_{\text{perf}}$                 & Bottom-3 queries                     & $\Delta_{\text{perf}}$                     \\
\midrule
\multirow{3}{*}{\textit{ramen}}          & sake cup                           & $+9.8$                                 & chopsticks                           & $+0.0$                                     \\
                                         & spoon handle                       & $+6.9$                                 & bowl                                 & $+0.5$                                     \\
                                         & wavy noodles                       & $+5.1$                                 & egg                                  & $+0.5$                                     \\
\midrule
\multirow{3}{*}{\textit{figurines}}      & rubber duck with hat               & $+18.1$                                & pumpkin                              & $-0.1$                                     \\
                                         & red apple                          & $+8.6$                                 & waldo                                & $+0.3$                                     \\
                                         & porcelain hand                     & $+4.5$                                 & pikachu                              & $+0.4$                                     \\
\midrule
\multirow{3}{*}{\textit{teatime}}        & bag of cookies                     & $+11.7$                                & plate                                & $-0.3$                                     \\
                                         & stuffed bear                       & $+7.0$                                 & sheep                                & $+0.1$                                     \\
                                         & bear nose                          & $+5.4$                                 & dall-e brand                         & $+0.5$                                     \\
\midrule
\multirow{3}{*}{\textit{waldo\_kitchen}} & Stainless steel pots               & $+10.7$                                & knife                                & $-0.5$                                     \\
                                         & pour-over vessel                   & $+7.2$                                 & ottolenghi                           & $+0.2$                                     \\
                                         & red cup                            & $+6.0$                                 & pot                                  & $+0.3$                                     \\
\bottomrule
\end{tabular}
\end{table}

\section{Performance with a weaker segmenter}
\label{sec:weak-segmenter}
Following LangSplat \cite{qin2023langsplat}, we use the Segment Anything model (SAM) \cite{kirillov2023segment} to extract class-agnostic segments for every image and use these segments to compute CLIP \cite{radford2021learning} embeddings for supervision (\cf \cref{eq:lang-gt}). To understand the effect of the choice of 2D segmenter on the final performance, we optimize our method as well as LangSplat using segments from a weaker model, namely Detic \cite{zhou2022detecting}. \Cref{tab:detic} shows that N2F2 (with Detic) performs slightly worse than N2F2 (with
SAM), but still outperforms LangSplat (with SAM).

\begin{table}[t]
\centering
\setlength{\tabcolsep}{6pt}
\caption{Performance comparison between our method and LangSplat~\cite{qin2023langsplat} using Detic~\cite{zhou2022detecting} as 2D segmenter.}
\label{tab:detic}
\begin{tabular}{lcc}
\toprule 
Method & \textit{figurines} & \textit{teatime} \\
\midrule
LangSplat w/ Detic & 43.2 & 63.9  \\
LangSplat w/ SAM (\textit{original}) & 44.7 & 65.1 \\
\hline
N2F2 w/ Detic & \textbf{45.5} & \textbf{66.2} \\
N2F2 w/ SAM (\textit{original}) & \textbf{47.0} & \textbf{69.2} \\
\bottomrule
\end{tabular}
\end{table}

\section{Experiments with scenes from ScanNet dataset}
\label{sec:scannet}
In this work, we mainly evaluate on the LERF and 3D-OVS datasets. To expand this evaluation, we also considering evaluating and comparing our method on the ScanNet dataset~\cite{dai2017scannet}. \Cref{tab:scannet} shows the performance comparison on 4 scenes: \textit{0050\_02}, \textit{0144\_01}, \textit{0300\_01} and \textit{0423\_02}.

\begin{table}[t]
\centering
\small
\setlength{\tabcolsep}{4pt}
\caption{Performance on ScanNet dataset.}
\label{tab:scannet}
\begin{tabular}{lcccc}
\toprule 
Method & \textit{0050\_02} & \textit{0144\_01} & \textit{0300\_01} & \textit{0423\_02}\\
\midrule
LangSplat & 60.2 & 56.5 & 54.7 & 57.0 \\
N2F2 & \textbf{68.1} & \textbf{61.7} & \textbf{59.9} & \textbf{60.4} \\
\bottomrule
\end{tabular}
\end{table}

\section{Analysis of backbone components}
\label{sec:backbone}
We analyze the different components of the N2F2 architecture and evaluate their impact on the final performance. Specifically, we compare using a pure-MLP backbone instead of the TriPlane+MLP backbone used in N2F2. We empirically observe that the MLP backbone is $2.3\times$ more memory-efficient than TriPlane+MLP. Notably, both these backbones use 7$\times$ and 3$\times$ less memory than LangSplat~\cite{qin2023langsplat} respectively.
Both backbones achieve similar final performance, but TriPlane+MLP converges 20$\times$ faster than the MLP backbone.

Additionally, we also experiment with removing the nested feature design in N2F2. We observe that this results in a drop of $2.0$ mIoU on segmentation and $2.5$\% on localization, still outperforming LangSplat.

\section{Open-vocabulary Retrieval Task}%
\label{sec:retrieve}

Next, we assess the performance of our method in the context of text-based retrieval.
The objective of the open-vocabulary retrieval task is to identify the most relevant object within the scene based on a natural language query.
Unlike segmentation, which assesses the precision of object boundaries, retrieval focuses more on the model's ability to find relevant objects.

\begin{figure}[t]
    \centering
    \includegraphics[width=\textwidth]{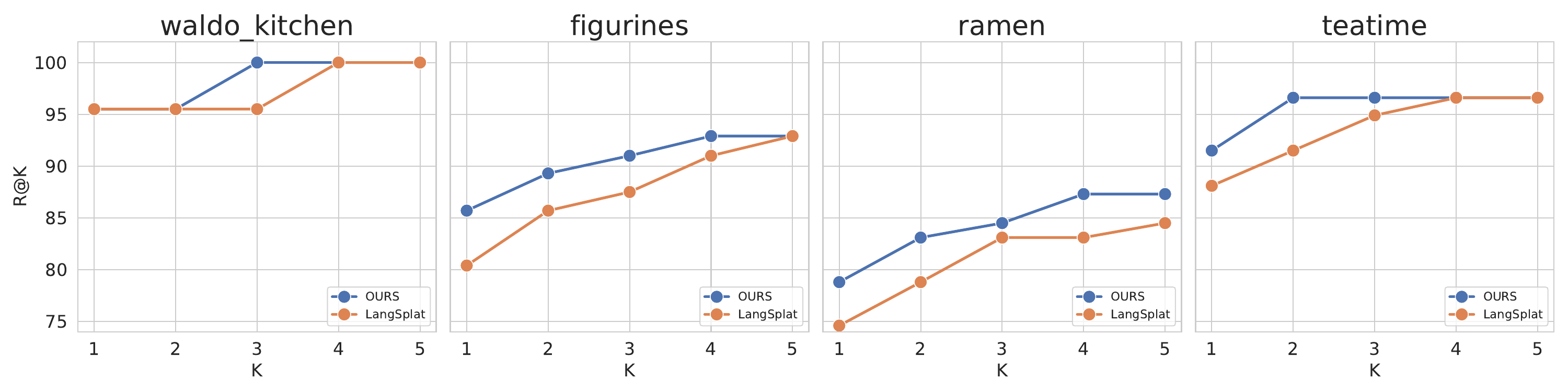}
    \caption{Open-vocabulary Retrieval performance on the expanded LERF dataset from Qin \etal~\cite{qin2023langsplat}.
    $R@K$ (\%) reported at $K=1,2,3,4,5$ for each scene.}%
    \label{fig:retrieval}
\end{figure}

To this end, we repurpose the expanded LERF dataset from Qin \etal~\cite{qin2023langsplat} for this task.
For a test view, we take the pool of segments from SAM as given,  compute a relevancy score \textit{per segment} (which is the average relevancy over segment pixels), and then rank these segments according to the predicted relevancy scores.
In this task, for a given text query, if the correct corresponding segment is within the top $K$ retrieved segments, then the query is said to be retrieved.
We report the recall ($R@K$) metric which is the fraction of correctly retrieved queries for a given $K$.
Note that the performance at $K=1$ is correlated with the localization accuracy of the method, indicating its ability to directly pinpoint the exact segment most relevant to the given text query.
\Cref{fig:retrieval} compares the performance of our method with LangSplat~\cite{qin2023langsplat} across a range of $K$.
We can see that at $K=1$, \method consistently performs better than LangSplat across scenes (which was also reflected in the localization performance reported in Table 1 of the main paper).
The $R@K$ of both methods increases with $K$, while the performance gap becomes smaller.

\section{Implementation Details}%
\label{sec:impl}

\subsection{Extracting Training data}%
\label{sec:gt}

To obtain embeddings for the text queries,
we leverage the OpenCLIP ViT-B/16 model with the segment masks obtained from the SAM ViT-H model.
We use the \texttt{laion2b\_s34b\_b88k} pretrained checkpoint for the OpenCLIP model.
For SAM mask generation, we sample pixels on a $32\times 32$ grid, set the minimum mask region area to be $100$ and use a IoU threshold of $0.7$.
To get an accurate geometry model for each scene, we first optimize the radiance field related parameters of the 3D Gaussian Splatting (3DGS) model using the RGB images for $30$k iterations.
We use the expected depth obtained from the 3DGS model to lift the SAM masks into 3D and obtain a physical scale for each mask as the largest eigenvalue of the covariance matrix of the lifted 3D segment-pixels.

\subsection{Feature Field Optimization}%
\label{sec:ffo}

Then, we optimize our feature field for another $30$k iterations while freezing all other parameters of the model.
As mentioned in the paper, the feature field is modeled with a TriPlane + 3-layer MLP\@.
We use a $512\times512$ resolution for each of the planes with $64$-dimensional features.
A hidden size of $256$ is used for the MLP\@.
We empirically observe that using a \textit{zero}-initialization for the projection matrix $\mathbf{W}$ gives the best results. The $\lambda$ weight used with the cosine loss (\cf Eq.\ (5)) is set to be $0.001$. We use a learning rate of $0.0016\times\text{scene\_extent}$ for the TriPlane and $0.00125$ for the MLP\@. All experiments and comparisons used an \textit{NVIDIA P40} (24GB RAM).
Our models are trained in $\approx$1 hour and take $\approx$600MB of memory.

\subsection{Rendering details}%
\label{sec:render}

As described in Sec.\ 3.1 of the paper, during rendering, we query the TriPlane+MLP representation at the Gaussian centers to get the associated feature for each Gaussian, and then use the Gaussian Splatting renderer to obtain the rendered feature map.
The default 3DGS implementation does not support rendering arbitrary features, so we modify and use the \textit{NDRasterizer} provided by Nerfstudio~\cite{nerfstudio} to render the features.

\subsection{Deferred rendering during training and not during testing}
\label{sec:deferred}
As described in Sec.\ 3.1 of the paper, we use deferred rendering to save memory during training, \ie we first render TriPlane features and then apply the MLP to obtain the pixel/ray feature. We do not do this during test time. This is because, as described in Sec.\ 3.3, we premultiply point features with the $\gamma^{3D}$ tensor to obtain the composite embedding. Hence, we need to maintain full-sized features per Gaussian during test-time.

Note that, applying the MLP on the rendered features (\ie MLP$(\alpha_1F_1+\alpha_2F_2+\dots)$) is not strictly equal to first applying the MLP on the point features and then rendering the full features (\ie $\alpha_1\text{MLP}(F_1)+\alpha_2\text{MLP}(F_2)+\dots$).
However, after training, Gaussians tend to be either transparent or opaque, so most pixels receive a non-negligible contribution from a single dominant Gaussian ($\alpha_i\approx 1$), making the train-time and test-time formulations approximately equivalent. 

\subsection{Efficient Scale-aware Feature Computation}%
\label{sec:scale-feat}

When optimizing the Hierarchical Loss (\cf Eq.\ (5)), different pixels $u$ in a batch will have different associated scales $s$ (and hence different mapped dimensions $M(s)$).
Thus, to efficiently compute the scale-aware features $\mathbf{W}_{1:M(s)}\theta(u)_{1:M(s)}$, we implement a batch-wise masking mechanism. Specifically, for a given scale $s$, we employ a binary mask $B(s)$ where entries corresponding to the active dimensions $1:M(s)$ are set to $1$, and all others are set to $0$.
The scale-aware features for each pixel $u$ are then computed as $\mathbf{W}_{1:M(s)}\theta(u)_{1:M(s)} = \mathbf{W} \cdot \left(B(s) \odot \theta(u)\right)$, where $\odot$ denotes element-wise multiplication.
The mask $B$ is computed batch-wise with negligible overhead.

\begin{table}[t]
    \centering
    \caption{Distinction between the \textit{simple} and \textit{compound} queries for each scene.}
    \begin{tabular}{l@{\hspace{25pt}}P{4.1cm}@{\hspace{25pt}}P{4.1cm}}
    \toprule
    Scene & Simple queries & compound queries \\
    \toprule
    \textit{waldo\_kitchen} & sink, refrigerator, cabinet, spatula, toaster, plate, ottolenghi, spoon, ketchup, pot, knife &
    yellow desk, Stainless steel pots, frog cup, red cup, pour-over vessel, plastic ladle, dark cup \\
    \midrule
    \textit{figurines} & jake, bag, spatula, porcelain hand, rubics cube, waldo, pumpkin, miffy, pirate hat, old camera, pikachu
    & tesla door handle, rubber duck with hat, red toy chair, toy elephant, green toy chair, pink ice cream, green apple, red apple, rubber duck buoy, toy cat statue \\
    \midrule
    \textit{ramen} & corn, plate, chopsticks, egg, bowl, kamaboko, onion segments, napkin, spoon, hand, nori
    & sake cup, glass of water, wavy noodles \\
    \midrule
    \textit{teatime} & three cookies, plate, hooves, apple, dall-e brand, coffee, sheep
    & bear nose, bag of cookies, tea in a glass, stuffed bear, coffee mug, paper napkin, yellow pouf \\
    \bottomrule
    \end{tabular}%
    \label{tab:comp_labels}
\end{table}

\section{Qualitative results}%
\label{sec:qual}

\begin{figure}
    \centering
    \includegraphics[width=\textwidth]{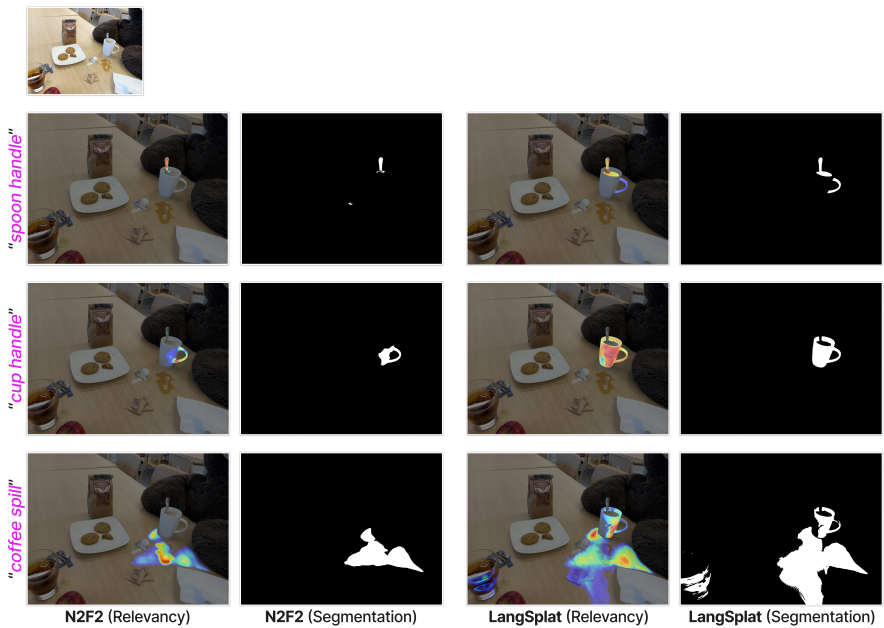}
    \caption{Scene: \textit{teatime}. Each row contains results for the text query shown on the \textbf{\textcolor{magenta}{left}}.
    \textbf{Columns}: N2F2 (Relevancy and Segmentation maps), LangSplat~\cite{qin2023langsplat} (Relevancy and Segmentation maps).}%
    \label{fig:teatime}
\end{figure}

\begin{figure}
    \centering
    \includegraphics[width=\textwidth]{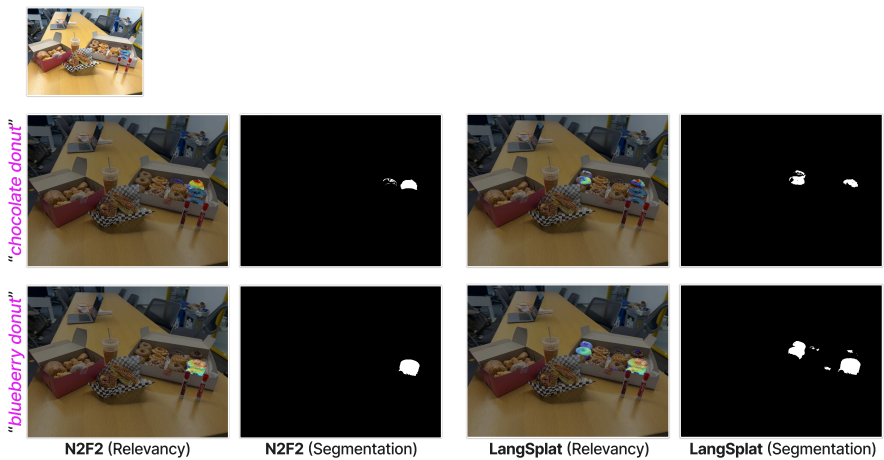}
    \caption{Scene: \textit{donuts}. Each row contains results for the text query shown on the \textbf{\textcolor{magenta}{left}}.
    \textbf{Columns}: N2F2 (Relevancy and Segmentation maps), LangSplat~\cite{qin2023langsplat} (Relevancy and Segmentation maps).}%
    \label{fig:donuts}
\end{figure}

\begin{figure}
    \centering
    \includegraphics[width=\textwidth]{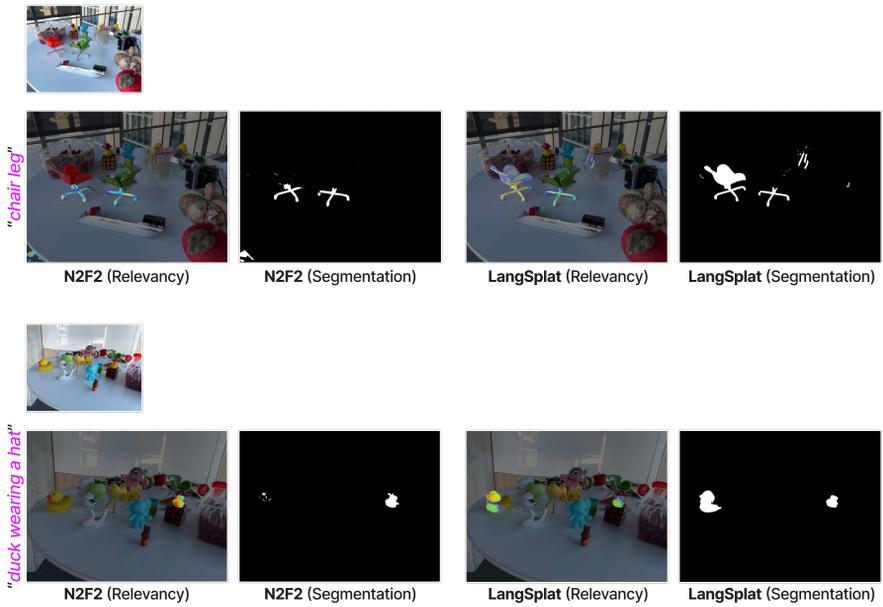}
    \caption{Scene: \textit{figurines}. Each row contains results for the text query shown on the \textbf{\textcolor{magenta}{left}}.
    \textbf{Columns}: N2F2 (Relevancy and Segmentation maps), LangSplat~\cite{qin2023langsplat} (Relevancy and Segmentation maps).}%
    \label{fig:figurines}
\end{figure}

In \cref{fig:teatime,fig:donuts,fig:figurines}, we show qualitative comparisons between \method and LangSplat~\cite{qin2023langsplat} on various challenging \textit{compounded} queries.

\end{document}